\documentclass[letterpaper, 10 pt, conference]{ieeeconf}  

\IEEEoverridecommandlockouts                              

\usepackage{graphicx} 
\usepackage{amsmath} 
\usepackage{amssymb}  
\usepackage{booktabs}

\title{\LARGE \bf
SLOT-V: Supervised Learning of Observer Models for Legible Robot Motion Planning in Manipulation
}

\author{Sebastian Wallkötter$^{1,2}$ and Mohamed Chetouani$^{3}$ and Ginevra Castellano$^{2}$
\thanks{$^{1}${\tt\small sebastian.wallkotter@it.uu.se}}%
\thanks{$^{2}$Department of Information Technology, Uppsala University, Uppsala, Sweden}%
\thanks{$^{3}$Institute for Intelligent Systems and Robotics, Sorbonne University, CNRS UMR 7222, Paris, France}%
}

\begin{document}

\maketitle
\thispagestyle{empty}
\pagestyle{empty}

\begin{abstract}
We present SLOT-V, a novel supervised learning framework that learns observer models (human preferences) from robot motion trajectories in a legibility context. Legibility measures how easily a (human) observer can infer the robot's goal from a robot motion trajectory. When generating such trajectories, existing planners often rely on an observer model that estimates the quality of trajectory candidates. These observer models are frequently hand-crafted or, occasionally, learned from demonstrations. Here, we propose to learn them in a supervised manner using the same data format that is frequently used during the evaluation of aforementioned approaches. We then demonstrate the generality of SLOT-V using a Franka Emika in a simulated manipulation environment. For this, we show that it can learn to closely predict various hand-crafted observer models, i.e., that SLOT-V's hypothesis space encompasses existing handcrafted models. Next, we showcase SLOT-V's ability to generalize by showing that a trained model continues to perform well in environments with unseen goal configurations and/or goal counts. Finally, we benchmark SLOT-V's sample efficiency (and performance) against an existing IRL approach and show that SLOT-V learns better observer models with less data. Combined, these results suggest that SLOT-V can learn viable observer models. Better observer models imply more legible trajectories, which may - in turn - lead to better and more transparent human-robot interaction.
\end{abstract}

\section{Introduction}
Transparency, a robot's ability to communicate any hidden internal state, is an element of artificial intelligence and robotics that is currently gaining in importance. For example, the European Union (EU) stated that transparency is an important factor to achieve trustworthy AI in its 2019 ethics guidelines \cite{euethics2019}. The IEEE, too, has recognized the need for transparency in autonomous systems and made a proposal towards its standardization \cite{winfield2021}. Further, several other ethical standards on the topic have stated the need for transparency \cite{winfield2019}.

At the same time, achieving transparency on a technical/implementation level is still a very active research topic.
In artificial intelligence and machine learning, one proposed answer is to use explainable AI (XAI) techniques, and several promising XAI approaches have been developed in recent years \cite{du2019}. Beyond XAI, robotics complements these techniques with domain-specific approaches that use a robot's embodiment or the situatedness of human-robot interaction scenarios \cite{wallkotter2021}.

One such approach to achieving transparency that is unique to robots goes by the name legibility \cite{dragan2013a,dragan2013b}. Legibility considers the scenario where a robot performs a goal-oriented movement under human supervision (referred to as the observer). In this scenario - which is typically a manipulation scenario -, the robot ought to alter its movement to communicate the intended goal and disambiguate it from alternative goals so that the observer - who is uncertain about the true goal - can quickly asses the situation. Computing such a legible trajectory requires specialized motion planning and several authors have suggested frameworks to accomplish this \cite{bodden2018,zhao2020,nikolaidis2016a}. A common trend among these frameworks is that they construct a mathematical model of the observer's expectations (the observer model) to vet candidate trajectories, which makes choosing a good observer model crucial to the framework success.

\begin{figure}[t]
    \centering
    \includegraphics[width=\linewidth]{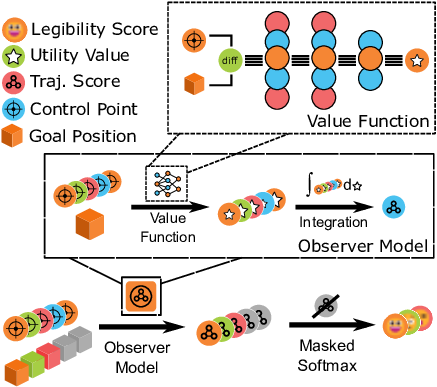}
    \caption{{\bf Schematic overview of SLOT-V.} (bottom:) SLOT-V computes the legibility of a trajectory by calling an observer model individually for each potential goal in the environment. (middle:) Assuming a (given) target goal, the observer model computes a trajectory score by applying a value function individually to each control point of a trajectory. (top:) The value function (here: a feed-forward neural network) estimates how legible it is to move through a given position to reach a given target goal. This is similar to the idea of a value function in RL (hence the name); however, SLOT-V is purely supervised.}
    \label{fig:arch-overview}
\end{figure}

Here, we take a closer look at the observer models used in legibility and address a commonly faced limitation of them, namely that they are frequently based on a researcher's intuition (hand-crafted). This hand-crafted nature of the observer model is usually described as either a limitation or a subject for future work, but - to our knowledge - still needs a satisfactory answer. Hence, we propose SLOT-V (fig. \ref{fig:arch-overview}), a novel supervised learning approach to extract the observer model from labeled robot motion trajectories and make the following contributions:
\begin{itemize}
    \item We present a novel framework (SLOT-V) that takes labeled robot trajectories as input and learns the observer model - a mathematical representation of the user's preferences regarding how the robot should move.
    \item We provide empirical evidence that (1) SLOT-V can learn a wide range of observer models, that (2) SLOT-V can generalize to unseen environments with different goal counts and/or configurations, and that (3) SLOT-V is more sample efficient than an alternative inverse reinforcement learning (IRL) approach.
\end{itemize}

\section{Related Work}
\subsection{Legibility}
One of the first major approaches to modelling legibility has been proposed by Dragan et al. \cite{dragan2013a,dragan2013b} and has sparked extensive follow up \cite{nikolaidis2016a, holladay2014, bied2020, dragan2015,chang2018,kebude2018,lemasurier2021}. The work assumes that humans expect robots to move efficiently and that we can model this expectation using a cost function over trajectories (the observer model). Using this function, we can then not only compute the most expected trajectory (called predictability), but can also compute a trajectory that maximizes the cost of moving towards alternative goals while still reaching the original target, i.e., a trajectory that minimizes the expectation that the robot moves to any alternative goal (called legibility). 

A follow-up to this idea has been proposed in Nikolaidis et al. \cite{nikolaidis2016a} under the title viewpoint-based legibility. Here legibility is not computed in world space. Instead the trajectory and any potential goals are first projected into a plane that is aligned with the observers point of view and then {\it Dragan legibility} is computed in the resulting space. This allows the robot to take into account the human's perspective and also allows it to account for occlusion from the perspective of the observer. Recently, this has been extended to multi-party interactions \cite{faria2021}. Further, a similar line of thinking has been used by Bodden et al. \cite{bodden2018}, where the authors project the trajectory into a goal-space, a (manually) designed latent space wherein it is easy to measure a trajectories expected goal. Then the authors compute the score of a trajectory as a sum of several terms of which one is the observer model which takes the form of an integral over the distance (measured in goal-space) between each point along the trajectory and the target goal.

What is interesting about the aforementioned approaches is that they all use an explicit observer model (a function that rates/scores the legibility of a given trajectory) and that they all engineer this observer model by hand (it is hand-crafted). Considering that the focus of the aforementioned papers is primarily on trajectory generation and motion-planning, hand-crafting the observer model is adequate, as the goal is to show the capabilities of the new planner. However, a hand-crafted model of human expectations might not be sophisticated enough to scale into complex environments, a challenge that is also recognized in aforementioned works as early as \cite{dragan2013b}. {\it Our contribution addresses this gap and suggests a new way to build such a (more sophisticated) observer model}.

Looking at recent works, Guletta et al. proposed HUMP \cite{gulletta2021}, a novel motion planner that generates human-like upper-limb movement, He et al. \cite{he2021} solved an inverse kinematics problem to create legible motion, Gabert et al. \cite{gabert2021} used a sampling-based motion planner to create obstacle avoidant yet legible trajectories, and Miura et al. extended the idea of legibility to stochastic environments \cite{miura2021a}. Looking at legibility more generally, the idea has also been explored in high-level task planning \cite{zhang2017} and other discrete domains \cite{bied2020,miura2021b,kulkarni2019a,macnally2018}. Other research explores legibility in the domain of navigation and this is getting more and more traction, likely due to a general rise of interest in autonomous driving. Here, a common framework is HAMP \cite{sisbot2007}, although several other methods exist \cite{calderita2021,bevins2021,che2020}. For more details, we recommend one of the several excellent reviews on the topic \cite{lichtenhaler2016, sreedharan2021,chakraborti2018b}.

\subsection{Machnine Learning in Legibility}
Shifting our attention towards data-driven approaches to legibility, we must first mention the system developed by Zhao et al. \cite{zhao2020}. The authors use learning from demonstration and inverse reinforcement learning to train a neural network that, from a partial manipulator trajectory, predicts the observer's expected goal. This model is then used to formulate a reward function that is used to create policies for legible movement using reinforcement learning. Lamb et al. \cite{lamb2017a,lamb2017b,lamb2018,lamb2019} assume that legible motion is human-like and use motion capture combined with model-identification to construct a controller that produces legible motion. Busch et al. \cite{busch2017} use direct policy search on a novel reward function that uses user feedback and asks the observer to guess the goal. Zhou et al. \cite{zhou2017} use a Bayesian model to learn timings when a robot should launch a movement. Zhang et al. \cite{zhang2017} and Beetz et al. \cite{beetz2010} independently explore learning strategies for high-level task planning, and, finally, Angelov et al. \cite{angelov2019} propose the interesting idea of using a causal model on the latent space of a deep auto-encoder to learn the task specifications that make movements legible.

Most of the above approaches have in common that they learn a policy from data and that this policy is later exploited to produce legible behavior. This is useful, because it allows us to learn important aspects of legibility directly from the observer, which is potentially more accurate than building such a policy by hand. A downside to learning a policy this way is that we loose the ability to easily transit to unseen environments. This works trivially for the planning methods introduced above, but requires retraining (and often new data) for policy-based methods. \textit{Our method does not suffer from this limitation (which we demonstrate), because we only learn the observer model and not a general model of the environment. This allows us to retain the important components of planning methods that enable this generalization}. This is similar to what Zhao et al. \cite{zhao2020} are doing, however, we do not learn from (optimal) demonstrations, which can be difficult to come by in a legibility context.

\section{Methods}
\subsection{SLOT-V}\label{sec:slot-v-details}
There are two insights behind SLOT-V: (1) Most legible planners improve trajectories using a gradient-based method on a trajectory's score, and we can not just compute a gradient with respect to the trajectory but also with respect to parameters in an observer model. (2) When evaluating planners, most authors generate sample trajectories, show these to users, and make them guess the robot's intended goal \cite{wallkotter2022}. The resulting frequency of guessing correctly can, in and off itself, be understood as a trajectory score and computing this score can be done on arbitrary trajectories and isn't limited to framework-optimal (result) trajectories only. As such, we can use this rating as a target score and build a system (SLOT-V) that can directly lean from it. Combined, this means that we can obtain legibility ratings from human feedback (emojis in figure \ref{fig:arch-overview}) and can then use these to learn an observer model's parameters in a supervised fashion using the same approach we'd otherwise use to optimize a trajectory (backprop wrt. model parameters instead of trajectory parameters).

Starting in the middle of figure \ref{fig:arch-overview}, we take a standard observer model $O:\Gamma\times\mathbb{G}\to\mathbb{R}^+$, which computes a score for a trajectory moving to a given goal, and allow it to have an arbitrary (differentiable) parametrization:

\begin{equation}
    O: \Gamma\times\mathbb{G}\times\Theta\to\mathbb{R}~,~(\gamma, \mathbf{g}, \mathbf{\theta})\mapsto\int_\gamma V(\mathbf{r}, \mathbf{g}, \mathbf{\theta})~\mathrm{d}t.
\end{equation}

Here $\gamma\in\Gamma$ is a trajectory from the set of all trajectories, $g\in\mathbb{G}$ is a goal from the finite set of potential goals, $V:~S\times\mathbb{G}\times\Theta \to \mathbb{R}^+$ is a value function that is parameterized by $\theta\in\Theta$ and defined on the elements of the planning space $S$ (which contains the set of trajectories $\Gamma$), $\mathbf{r}$ is a point in $\gamma$, and $t$ is an infinitesimal time step. We then apply this observer model to every element in $\mathbb{G}$ and stack the result into a score vector $\mathbf{s}$, which we define component-wise as
\begin{align}
    \mathcal{L}_\textrm{SLOT-V}: \Gamma\times\mathbb{G}^{|\mathbb{G}|}\times\Theta \to \mathbb{R}^{|\mathbb{G}|}~&,~ (\gamma, \mathbf{g}, \mathbf{\theta})\mapsto \mathbf{s}; \\
    &\mathbf{s}_i = O(\gamma,\mathbf{g_i},\theta).
\end{align}

The trajectory score vector $\mathbf{s}$ contains logits and higher values for $\mathbf{s}_i$ indicate higher odds that the user will think of the robot as moving to the goal $\mathbf{g}_i$. From these scores we can obtain a probability distribution by applying the softmax
\begin{equation}
    P(\mathbf{g}|\gamma,\theta) = \mathrm{Softmax}(\mathbf{\mathcal{L}_\textrm{SLOT-V}}).
\end{equation}

We can then pair this predicted distribution for a given trajectory $\gamma$ with a target distribution for that trajectory and learn the model parameters $\theta$ using a cross entropy loss. 

There are several ways to obtain such a target distribution. For example, we can ask users to guess the goal after having seen a trajectory and use the frequencies at which users pick potential goals to estimate this distribution. A technique, that is often used when evaluating novel planners that use hand-crafted observer models \cite{wallkotter2022}. Alternatively, we can obtain labels from existing legible planners by using them to compute the legibility score of a trajectory for each goal in the environment. We use the latter during our experiments to explore the hypothesis space of SLOT-V.

Shifting our attention to the value function $V$ (top element in figure \ref{fig:arch-overview}), we can understand it as a function that computes how legible it is to move through a given position $\mathbf{r}$ with the aim of ending in a specific goal $\mathbf{g}_i$. This is similar to the notion of a value function in reinforcement learning (how good is it to be in state $s$?), which is why we refer to $V$ as value function\footnote{Note, however, that we are not doing reinforcement learning framework here.} and choose to model it using a feed-forward neural network. One desirable property for this value function is that its resulting value doesn't change if we change the origin of the coordinate system in which $\mathbf{r}$ and $\mathbf{g}$ are expressed. We explicitly embed this ability into the model by adding a {\bf custom first layer} that expresses both inputs relative to the goal $\mathbf{g}$. This changes the input trajectory position $\mathbf{r}_\mathrm{in}$ to $\mathbf{r}_\mathrm{in} = \mathbf{r} - \mathbf{g}$ and the input goal position $\mathbf{g}_\mathrm{in}$ to $\mathbf{g}_\mathrm{in} = \mathbf{g} - \mathbf{g} = 0$ (which we omit). For the remaining layers, we use standard dense layers with ReLU \cite{nair2010} activations which we tune using a hyperparameter search with random rollouts as described in section \ref{sec:impl-details}.

\subsection{T-REX}
To gauge the utility of SLOT-V, we wanted to compare it against another algorithm that is capable of learning an observer model. For this, we adapted the inverse reinforcement learning algorithm called T-REX \cite{brown2019}. Compared to other IRL algorithms, T-REX has the ability to learn from sub-optimal demonstrations assuming we can define a ranking on pairs of demonstrations that indicates which demonstration was better. This allows us to train T-REX on the same dataset that we use to train SLOT-V because the legibility score implies such a ranking eliminating the confound of having different input data.

Similar to other IRL methods, T-REX learns a parameterized reward function $r: \Gamma\times\mathbb{G}\times\Theta\to\mathbb{R}$, which - in our context - predicts the reward a robot would obtain if it moved through a certain position $\mathbf{r}$ with the intent of reaching a target goal $\mathbf{g}$. Names aside, this is the same definition we used for SLOT-V's value function $V$ and, as such, we parameterize it using the same neural network archetype (but different hyperparameters). This eliminates further differences between the frameworks.

Using this reward function $r$ and parameterizing the trajectory $\gamma$ as a sequence of control points $\gamma^k$ (see section \ref{sec:environment}), we can compute the accumulated reward for a trajectory as the sum over estimated rewards at each control point 
\begin{equation}
R:\Gamma\times\mathbb{G}\times\Theta \to \mathbb{R}~,~ (\gamma,\mathbf{g},\theta) \mapsto \sum_{k=0}^N r(\gamma^k, \mathbf{g}, \theta).     
\end{equation}

If we now consider a pair of trajectories $\gamma_1$ and $\gamma_2$, then we can use our a priori knowledge of which one was more optimal (wlog. $\gamma_2$) to define a constraint on the accumulated reward
\begin{equation}
    R(\gamma_1, \mathbf{g}, \theta) < R(\gamma_2, \mathbf{g}, \theta).
\end{equation}

We can understand this constraint probabilistically and formulate a function that, given two trajectories, computes a likelihood of which trajectory is more optimal
\begin{align}
    \textrm{T-REX}: \Gamma\times\Gamma\times\mathbb{G}\times\Theta \to \mathbb{R}~,\\
    (\gamma_1, \gamma_2, \mathbf{g},\theta) \mapsto
    \mathrm{Softmax}\begin{pmatrix}
    R(\gamma_1, \mathbf{g}, \theta)\\
    R(\gamma_2, \mathbf{g}, \theta)
    \end{pmatrix}.
\end{align}
Using this function, and our a-priori ranking of demonstrations, we can define a binary classification problem (using, e.g., the second component $\textrm{T-REX}(\gamma_1,\gamma_2,\mathbf{g},\theta)_2$ as output) and we can learn optimal values for the parametrization $\theta$ using a binary crossentropy loss.

Once learned, we can use the reward function - in particular the accumulated reward $R$ - to once again compute a score for each potential goal in the environment
\begin{align}
    \mathcal{L}_\textrm{T-REX}: \Gamma\times\mathbb{G}^{|\mathbb{G}|}\times\Theta \to \mathbb{R}^{|\mathbb{G}|}~&,~ (\gamma, \mathbf{g}, \mathbf{\theta})\mapsto \mathbf{s}; \\
    &\mathbf{s}_i = R(\gamma,\mathbf{g_i},\theta),
\end{align}
which we once again interpret as indicators for how likely a goal is inferred from the trajectory $\gamma$ by the observer.

\subsection{Environment and Task}\label{sec:environment}
\begin{figure}[t]
    \centering
    \includegraphics[width=\linewidth]{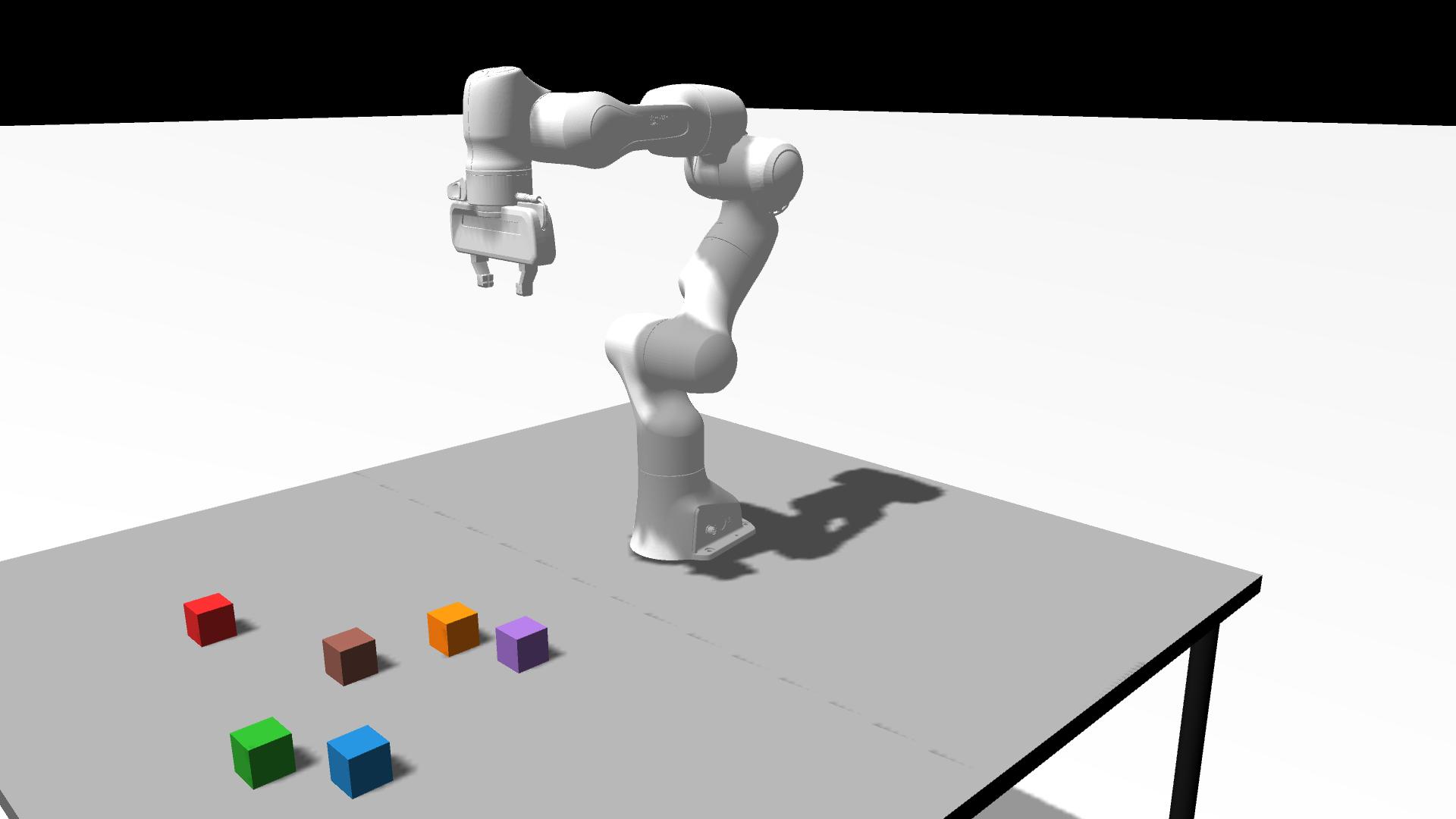}
    \caption{{\bf Example Environment} The render shows one of the environments used during trajectory generation. Different environments vary both the number and position of the cubes on the table.}
    \label{fig:env-sample}
\end{figure}

To situate the experiments and generate trajectories we use Ignition Gazebo\footnote{https://ignitionrobotics.org/home}, a state-of-the-art robot simulator, to create several pick-and-place environments from an environment template. In particular, we place a (desk-mounted) Franka Emika\footnote{https://www.franka.de/robot-system} in front of a second desk on which we randomly place several cubes. An example of such an environment is shown in figure \ref{fig:env-sample}, and to generate different environments, we vary both the number and position of cubes. 

We then task the robot with picking up one of the cubes in a legible manner and sample trajectories using the following steps: (1) Randomly choose a target goal. (2) Sample $N_\textrm{control}\in [3, 5]$ control points in the robot's workspace, which we limit to the area above the second table into which cubes are sampled. (3) Adding the starting position and the position of the chosen goal to the list of control points, create a piecewise-linear trajectory that moves through these points. (3) Resample the trajectory and place control points at regular intervals along it such that each trajectory has the same number of control points (here $100$).

\subsection{Experimental Procedure}\label{sec:procedure}
We aim to evaluate three properties of SLOT-V:
\begin{enumerate}
    \item We want to show that SLOT-V has the potential to replace existing hand-crafted observer models. For this, we aim to show that it can learn to imitate the models used in several existing legible planners, i.e., when trained on the output of a hand-crafted legibility model, SLOT-V will learn to make similar predictions as that model. 
    \item We want to show that SLOT-V (and our adapted implementation of T-REX) retain the ability to generalize to new, unseen environments. For this, we create multiple test sets ({\bf Trajectory}, {\bf Position}, and {\bf Goal Count}) that include previously unseen environments.
    \item We want to show that SLOT-V is more data efficient than the adapted T-REX implementation. For this, we assess the average performance that each method can achieve after having seen the same number of examples.
\end{enumerate}

To evaluate this, we thus proceed as follows: (1) We create a large dataset of randomly sampled trajectories by generating several environments and sampling trajectories therein. (2) For each sampled trajectory, we compute the legibility ratings of several existing legible planners, in particular the scores used by Dragan et al. \cite{dragan2013a} ({\it dragan legibility}), Nikolaidis et al. \cite{nikolaidis2016a} ({\it nikolaidis legibility}), and the two scores called EffDist and FastApp from Zhao et al. \cite{zhao2020} to provide target legibility ratings to learn (and hand-crafted models to imitate). (3) We perform hypertuning using random search (independently for SLOT-V and T-REX) to find good training and model hyperparameters. For this we train each model for 15 epochs (T-REX) or 5 epochs (SLOT-V - it converges much faster) and perform a total of 25 rollouts per framework. (4) Using the best set of hyperparameters for each framework, we train a final model for 25 (T-REX) and 15 (SLOT-V) epochs on each label set computed above and evaluate its performance using three test sets (described below). We repeat this process a total of $N=10$ times and report the mean result. (4) Finally, we train the best configuration of each framework on {\it Dragan legibility} for one epoch each and compute the validation accuracy every 10 updates. We do this in order to compare the data efficiency of both methods and we limit ourselves to a single epoch, because - at this point - the network has seen every example exactly once and further because the majority of the learning happens during the first few epochs. We repeat this process $10$ times and report averaged results.

\subsection{Implementation Details}\label{sec:impl-details}
\textit{Value/Reward Function} --- As mentioned above, we use the same simple neural network architecture to represent the value function in SLOT-V and the reward function in T-REX. Looking at the network's architecture, the networks input is first fed into the custom first layer described in section \ref{sec:slot-v-details}. We then add 1 (80\% of rollouts) or 2 (for 20\% of rollouts) dense hidden layers with $N\in\{1280, 1536, 1792\}$ units (depending on the rollout) and ReLU activation. We then add another dense layer with $N\in\{256, 512, 768, 1024\}$ units (again rollout dependent) and ReLU activation. We then finalize the network with a dense layer with $1$ unit and linear activation.

\textit{SLOT-V} --- As the training examples come from environments with different numbers of potential goals, we pad the set of potential goals $\mathbb{G}$ for environments with less than the maximum number of goals across environments (here $8$). For this, we insert dummy goals into the beginning of the pipeline (gray cubes in figure \ref{fig:arch-overview}), compute the legibility score $\mathcal{L}_\textrm{SLOT-V}$ for each goal including dummies and then remove/mask-out the inserted dummy goals before computing the softmax. This allows for better randomization of training data, because batches may now include examples from environments with different goal counts.

\textit{T-REX} --- To obtain training examples in the format required by T-REX, we first select two trajectories at random. Then, for each trajectory, we randomly select a target goal for which we wish to evaluate the accumulated reward and create the target label by comparing each trajectory's legibility score. During evaluation, we use the same goal padding approach we use for SLOT-V to enable efficient computation.

\subsection{Datasets}
We create a trajectory dataset using the following steps: (1) Using the environment template mentioned in section \ref{sec:environment}, create a set of $N_\textrm{env}$ environments with random goal positions goal counts taken from the set $\mathbb{G}$. (2) For each trajectory to generate, first (uniformly) choose a environment to sample in and then follow the trajectory generation steps outlined in section \ref{sec:environment}.

Using this method, we sampled a total of 7 dataset, one for training, three for validation, three for testing:
\begin{itemize}
    \item \textbf{Training} The training dataset contained a total of 100,000 trajectories sampled from $N_\textrm{env}=250$ environments using $\mathbb{G}_\textrm{train}=\{2, 3, 5, 6\}$ goals per environment. All models were trained on this data.
    \item \textbf{Trajectory}: One validation and one test set containing 10,000 trajectories sampled from the same 250 environments used to create the training set. (Note: validation and test set are sampled independently).
    \item \textbf{Position}: One validation and one test set containing 10,000 trajectories sampled from $10$ unseen environments (each) using $\mathbb{G_\textrm{train}}$. This set is used to estimate the model's ability to generalize across goal arrangements [same number of goals, but their configuration was unseen].
    \item \textbf{Goal Count}: One validation and one test set containing 10,000 trajectories sampled from $10$ unseen environments (different from {\bf Position}) using $\mathbb{G}_\textrm{val} = \{7\}$ or $\mathbb{G}_\textrm{test} = \{4, 8\}$ goals per environment. This set is used to estimate the model's ability to generalize over the total number of goals in the environment.
\end{itemize}

\section{Results}
\subsection{Hyperparameter Tuning}
{\it SLOT-V} --- Before starting the random search, we manually tuned SLOT-V until we found a choice of optimization hyperparameters that resulted in stable training. We then used the RMSprop optimizer with a learning rate chosen uniformly from $[.0004, .02]$, $\rho$ uniformly chosen from $[.7, .999]$, and momentum chosen uniformly from $[0, 1]$. For each trial, we also chose the batchsize randomly from $\{64, 128, 256, 512\}$.

After the planned $25$ rollouts, we noticed that none of the hyperparameters had a large effect on the performance with the exception of learning rate (lower performance for larger rates), and batchsize (less stable for larger batch sizes). We hence decided to compute another $25$ rollouts, reducing the batch size interval to $\{8, 16, 32, 64\}$.

From the results of these $50$ rollouts, we chose the following optimization hyperparameters: learning rate as $.005$, batch size as $32$, $\rho$ as $.9$ and momentum at $0$. Looking at the model hyperparameters, we chose to not use the extra hidden layer (no visible performance gain). We also chose $1536$ units for the first hidden layer and $768$ units for the second hidden layer.

{\it T-REX} --- Similar to SLOT-V, we started by manually tuning optimization hyperparameters until we found a configuration that resulted in stable training. For this, we had to switch from RMSprop to the Adam optimizer. We then optimized hyperparameters in a similar fashion, where we chose the learning rate from the interval $[1e-5, .001]$, $\beta_1 \in [.7, 1]$ and $\beta_2 \in [.7, 1]$. Batch sizes were chosen from the set $\{64, 128, 256\}$.

Larger learning rates achieved better results, and we chose a final rate of $.005$, outside the initially tested range. $\beta_1$ had no clear effect so we set it to the default of $.9$. Larger values of $\beta_2$ had a marginally negative effect on performance; however, due to human error we also keept it at the default value of $.999$. For the batch size, we chose $128$, because performance across levels was about equal, but variance was lowest for $128$. Looking at model parameters, we again chose to not use the additional hidden layer (no clear improvement). The first hidden layer benefitted from more units, so we chose to use $1792$ units for it. The second hidden layer showed a non-linear relationship with performance, so we used $768$ units as it was the best performing level in our search.

\subsection{Test Set Performance}
We trained and evaluated the models as described in section \ref{sec:procedure} and aggregated the results in table \ref{tab:legibility-scores}.

\begin{table}[t]
\caption{Accuracy of SLOT-V and T-REX on various Legibility Metrics}
\label{tab:legibility-scores}
\centering
\begin{tabular}{@{}llll@{}}
\toprule
\multicolumn{2}{r}{avg. of $10$ trails}    & SLOT-V & T-REX\\ 
 Metric     & Dataset & M ($\pm$ SD) & ($\pm$ SD) \\ \midrule
Dragan \cite{dragan2013a,dragan2013b}     & Training   & $0.94~(\pm~0.064)$   & $0.795~(\pm~0.182)$  \\
           & Trajectory & $0.938~(\pm~0.065)$  & $0.792~(\pm~0.18)$ \\
           & Goal Pos   & $0.895~(\pm~0.083)$  & $0.783~(\pm~0.17)$ \\
           & Goal Count & $0.922~(\pm~0.064)$  & $0.749~(\pm~0.198)$ \\
EffDist \cite{zhao2020}    & Training   & $1.0~(\pm~0.0)$      & $0.796~(\pm~0.032)$\\
           & Trajectory & $1.0~(\pm~0.0)$      & $0.798~(\pm~0.034)$\\
           & Goal Pos   & $1.0~(\pm~0.0)$      & $0.782~(\pm~0.033)$\\
           & Goal Count & $0.996~(\pm~0.005)$  & $0.772~(\pm~0.031)$\\
FastApp \cite{zhao2020}    & Training   & $0.973~(\pm~0.008)$  & $0.869~(\pm~0.034)$\\
           & Trajectory & $0.972~(\pm~0.008)$  & $0.866~(\pm~0.034)$\\
           & Goal Pos   & $0.972~(\pm~0.013)$  & $0.879~(\pm~0.024)$\\
           & Goal Count & $0.954~(\pm~0.017)$  & $0.86~(\pm~0.006)$\\
Nikolaidis \cite{nikolaidis2016a} & Training   & $0.911~(\pm~0.029)$  & $0.845~(\pm~0.033)$\\
           & Trajectory & $0.91~(\pm~0.031)$   & $0.844~(\pm~0.034)$\\
           & Goal Pos   & $0.875~(\pm~0.054)$  & $0.841~(\pm~0.039)$\\
           & Goal Count & $0.893~(\pm~0.041)$  & $0.796~(\pm~0.04)$\\ \bottomrule
\end{tabular}
\end{table}

Looking at the table, the first noticeable result is that SLOT-V shows excellent performance on the various test sets (hovering around 90\% accuracy). This indicates that SLOT-V successfully learned to imitate all observer models we tested here. Whats more is that we used the same architecture for each metric, which suggests that SLOT-Vs hypothesis space encompasses all these handcrafted metrics (and anything in-between). Looking at the individual metrics, we can notice that SLOT-V did worst when learning to imitate {\it Nikolaidis legibility} \cite{nikolaidis2016a}. On the other hand, it managed to almost perfectly imitate the {\it EffDist} metric.

Focusing on each metric individually, we can see that, as expected, test performance is highest for trajectories within previously seen environments ({\bf Trajectory}) and typically lowest for environments that had an unseen number of goals in them ({\bf Goal Count}). This is expected, because the environments in {\bf Goal Count} are not within the sample space of the {\bf Training} dataset, and are - because of that - hard to learn. What is interesting, however, is that both SLOT-V and T-REX still learn to perform well in these out-of-distribution environments, which we can see by how little the performance decreases in the test set compared to the training set.

Finally, we can see that SLOT-V consistently outperforms T-REX across all tested scenarios. This should not be missconstrued as T-REX being an inferior framework; in fact, it still performs very well on all metrics. Instead, it indicates that SLOT-V is better adapted to the current domain (learning observer models for legible motion generation), which is understandable, considering that it was designed for this very purpose.

\subsection{Sample Efficiency}
\begin{figure}[t]
    \centering
    \includegraphics[width=\linewidth]{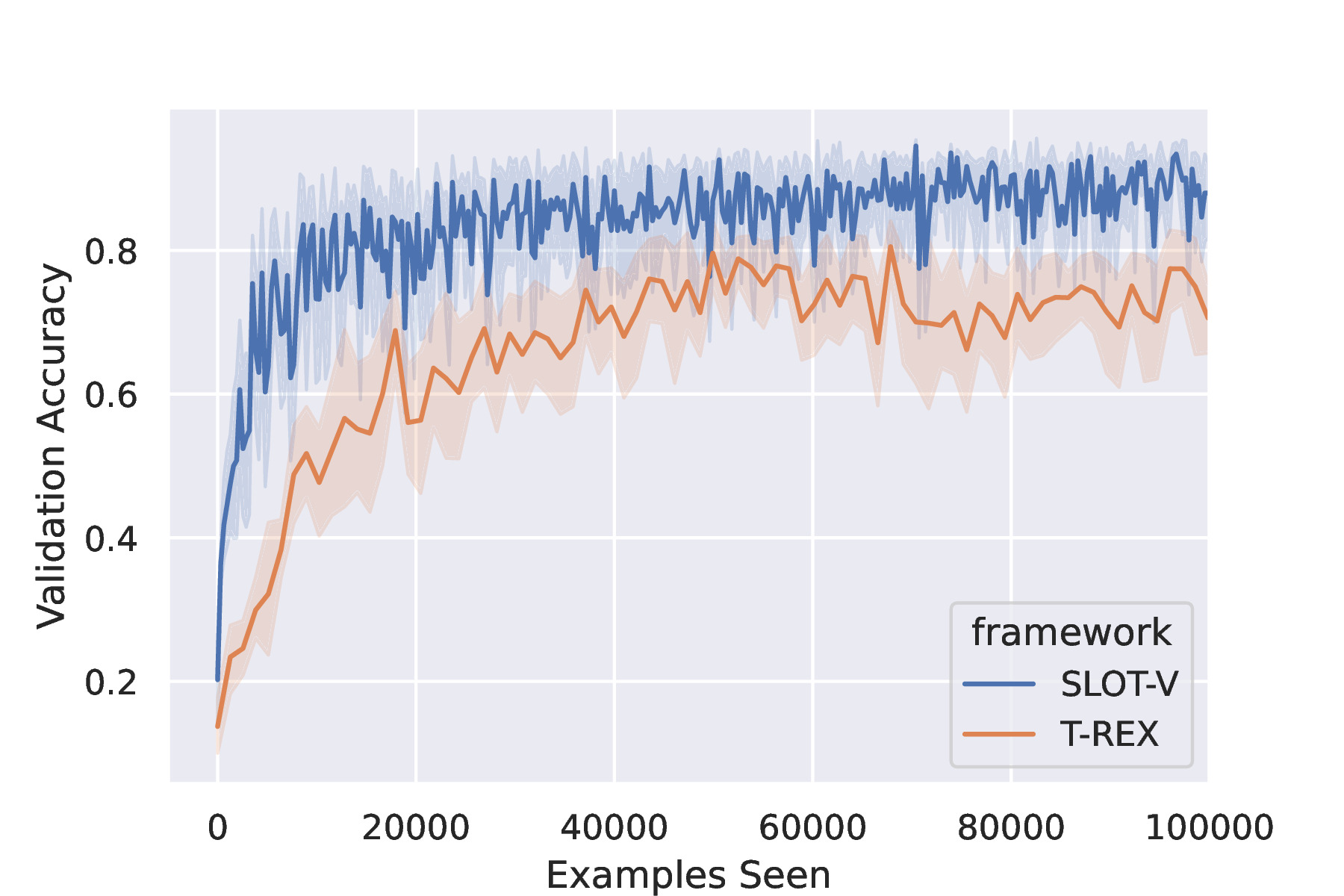}
    \caption{{\bf Sample Efficiency of SLOT-V and T-REX.} The graph shows the average ($N=10$) performance of both frameworks over the number of examples presented (higher is better). This is limited to the first epoch, because - at this point - the framework has seen every sample in the dataset exactly once. The gap between SLOT-V and T-REX indicates that SLOT-V is making more efficient usage of the data, meaning that it can learn better models when data is limited.}
    \label{fig:data-efficiency}
\end{figure}

Another aspect highlighting that SLOT-V is better adapted than T-REX is sample efficiency. The idea behind sample efficiency comes from the RL domain and measures how many samples are needed to obtain a desired level of performance. In our context, we can measure this by looking at the validation accuracy over the number of (unique) examples given to a framework, which we visualized in figure \ref{fig:data-efficiency}.

We can see that both frameworks learn at an exponential rate within the first 20,000 trajectories and continue to learn at a linear, but low rate thereafter. Further, we can see that SLOT-V - on top of learning the more performant model - learns much faster than T-REX, suggesting that it is the more sample efficient framework.

\section{Discussion}
Looking at the results, the most noticeable aspect is perhaps the ability of SLOT-V and T-REX to generalize to unseen environments, including out-of-distribution ones. We think that there are two reasons behind this. 

The first is probably the custom layer that we introduced in the value/reward function, which shifts the origin of the coordinate frame into the goal's position. Adding this layer is, in some sense, a form of feature engineering and incorporates expert knowledge into the model, because it allows the network to focus on the relative position of the robot and the goal rather than having discover the importance of this relationship. While this could backfire in other contexts, e.g., in cases where there is no natural choice for the coordinate frame's origin, this does not seem to be the case here and - as the results indicate - works well.

The second reason is that we only allow the observer model to depend on a single goal instead of all goals in the environment making the model independent of the total number of goals present. Approaches with hand-crafted observer models tend to follow this approach, whereas frameworks that rely on learning a policy use all available goals to compute a score. We think that this difference is one of the main drivers of the strong generalization capacity of existing hand-crafted planners, as this makes them easy to apply to new environments. Hence, we designed SLOT-V to retain this ability.

A second noteworthy aspect of our results is that SLOT-V achieved high performance on {\it Nikolaidis legibility} \cite{nikolaidis2016a}. This is interesting, because {\it Nikolaidis legibility} is based on {\it Dragan legibility}, but - contrary to it - applies the metric in a (perspective-)projected 2D space instead of planning in 3D world space. As such, SLOT-V had to learn how to represent this projection on top of learning to imitate {\it Dragan legibility}. Despite this challenge, performance is only slightly reduced in comparison, which we interpret as a testimony of SLOT-Vs capabilities.

Looking at our results on sample efficiency, we can see that SLOT-V is an improvement in this area as well. While it, like other neural-network based approaches, remains a rather data-hungry technique, we think that it is non-the-less a good step in the direction of finding techniques that bring the power of neural-networks into the human-robot interaction domain where only limited data is available.

\section{Limitations}
One limitation of this paper is that we only trained on ratings obtained from existing legible planners instead of training on human-labeled trajectories. This allows us to establish the efficacy that SLOT-V can learn the observer model - we would expect SLOT-V to be at least as good as existing hand-crafted models -, but doesn't explore the full limits of its capability. The reason why we didn't explore human labeled data is because we wanted to first establish this general efficacy and further gauge how much data we would have to collect to tackle the problem of learning a model from human data. Addressing this limitation represents a clear next step and future work for this research.

The second limitation is, in fact, the amount of data needed to train a good observer model, which is a limitation of SLOT-V itself and a limitation of any machine learning approach in general. While SLOT-V does show improved data efficiency compared to general IRL approaches like T-REX, it still required about $10,000$ unique examples to achieve 80\% validation accuracy. Clearly, this is still too high a number for humans to annotate manually; however, it is still an improvement over alternatives (T-REX requires $10$x more data), and the number might be exaggerated, because - during actual training - we can show examples several times. In the future, it would be interesting to see if this number can be reduced to a more managable amount, e.g. by more intelligently sampling trajectories.

A third, context dependent limitation may be the resulting model’s explainability and (provable) safety. Existing hand-crafted approaches provide a fully analytic model, which makes them explainable by design and may allow us to derive analytical safety guarantees. Neural network based approaches, like the one we chose here, are currently renowned to lack explainability or the ability to prove safe control. For such cases where safety is a concern we can however still use SLOT-V. Instead of using a neural network as model archetype we can use an explicit analytic model for which we only learn certain parameter values. This would enable us to again provide safety guarantees while also benefiting from learning-based adaptation.

\section{Conclusion}
Above we present SLOT-V a {\bf s}upervised {\bf l}earning framework to extract {\bf o}bserver models from {\bf t}rajectory data. It sets itself apart from other learning-based legibility methods in that it only learns the observer model, instead of learning a policy that is specific to the environment. We show that this allows for generalization not just across different trajectories in the same environment but also across different environments including completely unseen (out-of-distribution) ones. We also demonstrate SLOT-Vs ability to imitate several existing hand-crafted observer models, suggesting that it can be a viable replacement of them. We then compare SLOT-V's sample efficiency to the sample efficiency of T-REX, a state-of-the-art inverse reinforcement learning algorithm and show that - in the studied domain - SLOT-V learns better models from less data.

Combined, we think that these results show that SLOT-V is capable of learning useful observer models and hope that it can become a stepping stone towards more transparent and intuitive human-robot interaction.

\addtolength{\textheight}{-3cm}   




\section*{ACKNOWLEDGMENT}
This project has received funding from the European Union's Horizon 2020 research and innovation programme under grant agreement No 765955.


\bibliographystyle{IEEEtran}
\bibliography{references}

\begin{thebibliography}{10}
\providecommand{\url}[1]{#1}
\csname url@samestyle\endcsname
\providecommand{\newblock}{\relax}
\providecommand{\bibinfo}[2]{#2}
\providecommand{\BIBentrySTDinterwordspacing}{\spaceskip=0pt\relax}
\providecommand{\BIBentryALTinterwordstretchfactor}{4}
\providecommand{\BIBentryALTinterwordspacing}{\spaceskip=\fontdimen2\font plus
\BIBentryALTinterwordstretchfactor\fontdimen3\font minus
  \fontdimen4\font\relax}
\providecommand{\BIBforeignlanguage}[2]{{%
\expandafter\ifx\csname l@#1\endcsname\relax
\typeout{** WARNING: IEEEtran.bst: No hyphenation pattern has been}%
\typeout{** loaded for the language `#1'. Using the pattern for}%
\typeout{** the default language instead.}%
\else
\language=\csname l@#1\endcsname
\fi
#2}}
\providecommand{\BIBdecl}{\relax}
\BIBdecl

\bibitem{euethics2019}
H.-L. E.~G. on~Artificial~Intelligence, A.~HLEG, and H.-L. E.~G.
  on~Artificial~Intelligence, ``Ethics guidelines for trustworthy ai,'' p.~41,
  2019.

\bibitem{winfield2021}
A.~F. Winfield, S.~Booth, L.~A. Dennis, T.~Egawa, H.~Hastie, N.~Jacobs, R.~I.
  Muttram, J.~I. Olszewska, F.~Rajabiyazdi, A.~Theodorou \emph{et~al.}, ``Ieee
  p7001: a proposed standard on transparency,'' \emph{Frontiers in Robotics and
  AI}, p. 225, 2021.

\bibitem{winfield2019}
A.~Winfield, ``Ethical standards in robotics and ai,'' \emph{Nature
  Electronics}, vol.~2, no.~2, pp. 46--48, 2019.

\bibitem{du2019}
M.~Du, N.~Liu, and X.~Hu, ``Techniques for interpretable machine learning,''
  \emph{Communications of the ACM}, vol.~63, no.~1, pp. 68--77, 2019.

\bibitem{wallkotter2021}
S.~Wallk{\"o}tter, S.~Tulli, G.~Castellano, A.~Paiva, and M.~Chetouani,
  ``Explainable embodied agents through social cues: a review,'' \emph{ACM
  Transactions on Human-Robot Interaction (THRI)}, vol.~10, no.~3, pp. 1--24,
  2021.

\bibitem{dragan2013a}
A.~Dragan and S.~Srinivasa, ``{Generating Legible Motion}.''\hskip 1em plus
  0.5em minus 0.4em\relax none, 2013, p.~8.

\bibitem{dragan2013b}
A.~D. Dragan, K.~C. Lee, and S.~S. Srinivasa, ``{Legibility and Predictability
  of Robot Motion},'' in \emph{7th ACM/IEEE International Conference on
  Human-Robot Interaction (HRI)}.\hskip 1em plus 0.5em minus 0.4em\relax Tokyo:
  IEEE, 2013, pp. 301--308.

\bibitem{bodden2018}
C.~Bodden, D.~Rakita, B.~Mutlu, and M.~Gleicher, ``{A flexible
  optimization-based method for synthesizing intent-expressive robot arm
  motion},'' \emph{International Journal of Robotics Research}, vol.~37,
  no.~11, pp. 1376--1394, sep 2018.

\bibitem{zhao2020}
X.~Zhao, T.~Fan, D.~Wang, Z.~Hu, T.~Han, and J.~Pan, ``{An Actor-Critic
  Approach for Legible Robot Motion Planner},'' in \emph{Proceedings - IEEE
  International Conference on Robotics and Automation}, 2020.

\bibitem{nikolaidis2016a}
S.~Nikolaidis, A.~Dragan, and S.~Srinivasa, ``{Viewpoint-based legibility
  optimization},'' in \emph{ACM/IEEE International Conference on Human-Robot
  Interaction}, vol. 2016-April, 2016, pp. 271--278.

\bibitem{holladay2014}
R.~M. Holladay, A.~D. Dragan, and S.~S. Srinivasa, ``Legible robot pointing,''
  in \emph{The 23rd IEEE International Symposium on robot and human interactive
  communication}.\hskip 1em plus 0.5em minus 0.4em\relax IEEE, 2014, pp.
  217--223.

\bibitem{bied2020}
M.~Bied and M.~Chetouani, ``{Exploring the difference between solving and
  teaching in sensorimotor tasks},'' in \emph{ACM/IEEE International Conference
  on Human-Robot Interaction}, 2020, pp. 139--141.

\bibitem{dragan2015}
A.~D. Dragan, S.~Bauman, J.~Forlizzi, and S.~S. Srinivasa, ``{Effects of Robot
  Motion on Human-Robot Collaboration},'' in \emph{ACM/IEEE International
  Conference on Human-Robot Interaction}, vol. 2015-March, 2015, pp. 51--58.

\bibitem{chang2018}
M.~{Lee Chang}, R.~A. Gutierrez, P.~Khante, E.~{Schaertl Short}, and
  A.~{Lockerd Thomaz}, ``{Effects of Integrated Intent Recognition and
  Communication on Human-Robot Collaboration},'' in \emph{IEEE International
  Conference on Intelligent Robots and Systems}, 2018, pp. 3381--3386.

\bibitem{kebude2018}
D.~Keb{\"{u}}de, C.~Eteke, T.~M. Sezgin, and B.~Akg{\"{u}}n, ``{Communicative
  cues for reach-to-grasp motions: From humans to robots: Robotics Track},'' in
  \emph{Proceedings of the International Joint Conference on Autonomous Agents
  and Multiagent Systems, AAMAS}, vol.~2.\hskip 1em plus 0.5em minus
  0.4em\relax Stockholm, Sweden: ACM, 2018, pp. 874--882.

\bibitem{lemasurier2021}
G.~Lemasurier, G.~Bejerano, V.~Albanese, J.~Parrillo, H.~A. Yanco, N.~Amerson,
  R.~Hetrick, and E.~Phillips, ``Methods for expressing robot intent for
  human--robot collaboration in shared workspaces,'' \emph{ACM Transactions on
  Human-Robot Interaction (THRI)}, vol.~10, no.~4, pp. 1--27, 2021.

\bibitem{faria2021}
M.~Faria, F.~S. Melo, and A.~Paiva, ``Understanding robots: Making robots more
  legible in multi-party interactions,'' in \emph{2021 30th IEEE International
  Conference on Robot \& Human Interactive Communication (RO-MAN)}.\hskip 1em
  plus 0.5em minus 0.4em\relax IEEE, 2021, pp. 1031--1036.

\bibitem{gulletta2021}
G.~Gulletta, E.~C.~e. Silva, W.~Erlhagen, R.~Meulenbroek, M.~F.~P. Costa, and
  E.~Bicho, ``{A Human-like Upper-limb Motion Planner: Generating naturalistic
  movements for humanoid robots},'' \emph{International Journal of Advanced
  Robotic Systems}, vol.~18, no.~2, 2021.

\bibitem{he2021}
C.~He, X.-W. Xu, X.-F. Zheng, C.-H. Xiong, Q.-L. Li, W.-B. Chen, and B.-Y. Sun,
  ``Anthropomorphic reaching movement generating method for human-like upper
  limb robot,'' \emph{IEEE Transactions on Cybernetics}, 2021.

\bibitem{gabert2021}
C.~G{\"a}bert, S.~Kaden, and U.~Thomas, ``Generation of human-like arm motions
  using sampling-based motion planning,'' in \emph{2021 IEEE/RSJ International
  Conference on Intelligent Robots and Systems (IROS)}.\hskip 1em plus 0.5em
  minus 0.4em\relax IEEE, pp. 2534--2541.

\bibitem{miura2021a}
S.~Miura, A.~L. Cohen, and S.~Zilberstein, ``Maximizing legibility in
  stochastic environments,'' in \emph{2021 30th IEEE International Conference
  on Robot \& Human Interactive Communication (RO-MAN)}.\hskip 1em plus 0.5em
  minus 0.4em\relax IEEE, 2021, pp. 1053--1059.

\bibitem{zhang2017}
Y.~L. Zhang, S.~Sreedharan, A.~Kulkarni, T.~Chakraborti, H.~H.~H. Zhuo, and
  S.~Kambhampati, ``{Plan explicability and predictability for robot task
  planning},'' \emph{2017 IEEE International Conference on Robotics and
  Automation (ICRA)}, pp. 1313--1320, 2017.

\bibitem{miura2021b}
S.~Miura and S.~Zilberstein, ``A unifying framework for observer-aware planning
  and its complexity,'' in \emph{Uncertainty in Artificial Intelligence}.\hskip
  1em plus 0.5em minus 0.4em\relax PMLR, 2021, pp. 610--620.

\bibitem{kulkarni2019a}
\BIBentryALTinterwordspacing
A.~Kulkarni, S.~Srivastava, and S.~Kambhampati, ``{A Unified Framework for
  Planning in Adversarial and Cooperative Environments},'' \emph{Proceedings of
  the AAAI Conference on Artificial Intelligence}, vol.~33, pp. 2479--2487,
  2019. [Online]. Available: \url{www.aaai.org}
\BIBentrySTDinterwordspacing

\bibitem{macnally2018}
A.~M. MacNally, N.~Lipovetzky, M.~Ramirez, and A.~R. Pearce, ``{Action
  selection for transparent planning},'' in \emph{Proceedings of the
  International Joint Conference on Autonomous Agents and Multiagent Systems,
  AAMAS}, vol.~2, 2018, pp. 1327--1335.

\bibitem{sisbot2007}
E.~A. Sisbot, K.~F. Marin-Urias, R.~Alami, and T.~Sim{\'{e}}on, ``{A human
  aware mobile robot motion planner},'' in \emph{IEEE Transactions on
  Robotics}, vol.~23, 2007, pp. 874--883.

\bibitem{calderita2021}
L.~Calderita, A.~Vega, P.~Bustos, and P.~N{\'u}{\~n}ez, ``A new human-aware
  robot navigation framework based on time-dependent social interaction spaces:
  An application to assistive robots in caregiving centers,'' \emph{Robotics
  and Autonomous Systems}, vol. 145, p. 103873, 2021.

\bibitem{bevins2021}
A.~Bevins and B.~A. Duncan, ``Aerial flight paths for communication: How
  participants perceive and intend to respond to drone movements,'' in
  \emph{Proceedings of the 2021 ACM/IEEE International Conference on
  Human-Robot Interaction}, 2021, pp. 16--23.

\bibitem{che2020}
Y.~Che, A.~M. Okamura, and D.~Sadigh, ``Efficient and trustworthy social
  navigation via explicit and implicit robot--human communication,'' \emph{IEEE
  Transactions on Robotics}, vol.~36, no.~3, pp. 692--707, 2020.

\bibitem{lichtenhaler2016}
C.~Lichtenth{\"{a}}ler and A.~Kirsch, ``{Legibility of Robot Behavior : A
  Literature Review},'' \emph{HAL-archives}, vol.~1, no.~1, pp. 1--22, apr
  2016.

\bibitem{sreedharan2021}
S.~Sreedharan, A.~Kulkarni, D.~Smith, and S.~Kambhampati, ``A unifying bayesian
  formulation of measures of interpretability in human-ai interaction,'' in
  \emph{International Joint Conference on Artificial Intelligence}, 2021, pp.
  4602--4610.

\bibitem{chakraborti2018b}
T.~Chakraborti, A.~Kulkarni, S.~Sreedharan, D.~E. Smith, and S.~Kambhampati,
  ``{Explicability? Legibility? Predictability? Transparency? Privacy?
  Security? The Emerging Landscape of Interpretable Agent Behavior},'' in
  \emph{arXiv: 1811.09722}.\hskip 1em plus 0.5em minus 0.4em\relax unknown:
  ArXiv, 2018.

\bibitem{lamb2017a}
M.~Lamb, T.~Lorenz, S.~Harrison, R.~Kallen, A.~Minai, and M.~Richardson,
  ``{Behavioral Dynamics and Action Selection in a Joint Action Pick-and-Place
  Task},'' in \emph{Proceedings of the Annual Meeting of the Cognitive Science
  Society (CogSci)}, vol.~1.\hskip 1em plus 0.5em minus 0.4em\relax CogSci,
  2017, pp. 2506--2511.

\bibitem{lamb2017b}
M.~Lamb, R.~W. Kallen, S.~J. Harrison, M.~{Di Bernardo}, A.~Minai, and M.~J.
  Richardson, ``{To pass or not to pass: Modeling the movement and affordance
  dynamics of a pick and place task},'' \emph{Frontiers in Psychology}, vol.~8,
  no. JUN, jun 2017.

\bibitem{lamb2018}
M.~Lamb, R.~Mayr, T.~Lorenz, A.~A. Minai, and M.~J. Richardson, ``{The Paths We
  Pick Together: A Behavioral Dynamics Algorithm for an HRI Pick-and-Place
  Task},'' in \emph{Proceedings of the 2018 ACM/IEEE International Conference
  on Human-Robot Interaction}.\hskip 1em plus 0.5em minus 0.4em\relax Chicago:
  IEEE, mar 2018, pp. 165--166.

\bibitem{lamb2019}
M.~Lamb, P.~Nalepka, R.~W. Kallen, T.~Lorenz, S.~J. Harrison, A.~A. Minai, and
  M.~J. Richardson, ``{A Hierarchical Behavioral Dynamic Approach for Naturally
  Adaptive Human-Agent Pick-and-Place Interactions},'' \emph{Complexity}, vol.
  2019, 2019.

\bibitem{busch2017}
B.~Busch, J.~Grizou, M.~Lopes, and F.~Stulp, ``{Learning Legible Motion from
  Human–Robot Interactions},'' \emph{International Journal of Social
  Robotics}, vol.~9, no.~5, pp. 765--779, nov 2017.

\bibitem{zhou2017}
A.~Zhou, D.~Hadfield-Menell, A.~Nagabandi, and A.~D. Dragan, ``{Expressive
  Robot Motion Timing},'' in \emph{ACM/IEEE International Conference on
  Human-Robot Interaction}, vol. Part F1271.\hskip 1em plus 0.5em minus
  0.4em\relax New York, New York, USA: IEEE, mar 2017, pp. 22--31.

\bibitem{beetz2010}
M.~Beetz, F.~Stulp, P.~Esden-Tempski, A.~Fedrizzi, U.~Klank, I.~Kresse,
  A.~Maldonado, and F.~Ruiz, ``{Generality and legibility in mobile
  manipulation: Learning skills for routine tasks},'' \emph{Autonomous Robots},
  vol.~28, no.~1, pp. 21--44, jan 2010.

\bibitem{angelov2019}
D.~Angelov, Y.~Hristov, and S.~Ramamoorthy, ``Using causal analysis to learn
  specifications from task demonstrations,'' \emph{arXiv preprint
  arXiv:1903.01267}, 2019.

\bibitem{wallkotter2022}
S.~Wallkotter, M.~Chetouani, and G.~Castellano, ``A new approach to evaluating
  legibility: Comparing legibility frameworks using framework-independent robot
  motion trajectories,'' \emph{arXiv preprint arXiv:2201.05765}, 2022.

\bibitem{nair2010}
V.~Nair and G.~E. Hinton, ``Rectified linear units improve restricted boltzmann
  machines,'' in \emph{Icml}, 2010.

\bibitem{brown2019}
D.~Brown, W.~Goo, P.~Nagarajan, and S.~Niekum, ``Extrapolating beyond
  suboptimal demonstrations via inverse reinforcement learning from
  observations,'' in \emph{International conference on machine learning}.\hskip
  1em plus 0.5em minus 0.4em\relax PMLR, 2019, pp. 783--792.

\end{thebibliography}

\end{document}